\documentclass[11pt,a4paper]{article}
\usepackage[hyperref]{acl2017}
\usepackage{times}
\usepackage{latexsym}
\usepackage{tikz}
\usepackage{CJK}
\usepackage{amsmath}
\usepackage{amssymb}
\DeclareMathOperator*{\argmax}{argmax}
\usetikzlibrary{fit,positioning,arrows.meta}
\usepackage{float}
\usepackage{tabularx}

\usepackage{multirow}

\usepackage{hyperref}

\usepackage{url}

\aclfinalcopy 

\title{Cross-lingual Word Segmentation and Morpheme Segmentation as Sequence Labelling}

\author{Yan Shao \\
  Department of Linguistics and Philology, Uppsala University \\
  {\tt yan.shao@lingfil.uu.se}}

\date{}

\begin{document}
\maketitle
\begin{abstract}
This paper presents our segmentation system developed for the MLP 2017 shared tasks on
cross-lingual word segmentation and morpheme segmentation. We model both word and morpheme segmentation as character-level sequence labelling tasks. The prevalent bidirectional recurrent neural network with conditional random fields as the output interface is adapted as the baseline system, which is further improved via ensemble decoding. Our universal system is applied to and extensively evaluated on all the official data sets without any language-specific adjustment. The official evaluation results indicate that the proposed model achieves outstanding accuracies both for word and morpheme segmentation on all the languages in various types when compared to the other participating systems.  
\end{abstract}

\section{Introduction}

In natural language processing, word segmentation and morpheme segmentation are the initial steps to identify basic linguistic units, namely words and morphemes, for further analysis in higher-level tasks. Word segmentation can be very non-trivial, especially for languages without explicit indicators for word boundaries, such as Chinese, Japanese and Vietnamese. For morphologically rich languages like Turkish, words are further segmented into morphemes, such as stems, prefixes and suffixes for morphological analysis. Similar to non-trivial word segmentation, there are no clear boundaries between morphemes in the surface forms of words. Both word segmentation and morpheme segmentation can be viewed as identifying valid boundaries between consecutive characters.

Word segmentation is often formalised as a character-based sequence labelling problem to predict position tags \cite{xue2003chinese,udpipe}. Standard machine algorithms, such as Maximum Entropy \cite{berger1996maximum,low2005maximum}, Conditional Random Fields (CRF) \cite{lafferty2001crf,peng2004chinese} and neural networks \cite{chen2015long} are applied for the task in previous research.  Additionally, a number of word-based approaches have also been proposed \cite{zhang2007chinese,cai2016neural}. 

For morpheme segmentation, apart from unsupervised methods \cite{creutz2007unsupervised,poon2009unsupervised}, \newcite{ruokolainen2013supervised} model the task as sequence labelling, similarly to character-based word segmentation. They use CRF to predict position tags given words as sequences of characters. Furthermore, instead of employing traditional statistical models, \newcite{wang2016mor} propose and apply several recurrent neural network architectures to avoid heavy feature engineering. 

Considering the similarities between word segmentation and morpheme segmentation, we present a universal neural sequence labelling model that is capable of solving both segmentation tasks in this paper. Our baseline model is an adaptation of a bidirectional recurrent neural network (RNN) using conditional random fields (CRF) as the output interface for sentence-level optimisation (BiRNN-CRF). BiRNN-CRF achieves state-of-the-art accuracies on various sequence labelling tasks \cite{huang2015bidirectional, maend}. We modify the conventional position tags used for word segmentation so that they are also applicable to morpheme segmentation. Furthermore, a simple ensemble decoding technique is implemented to obtain additional improvements over the baseline model. Our system is fully data-driven and language-independent. It is extensively evaluated on the MLP 2017 shared task data sets. 

\section{Segmentation Model}

\subsection{Baseline Model}

\begin{CJK}{UTF8}{gbsn}
\begin{figure}
\scalebox{0.78}{
\begin{tikzpicture}[x=1.5cm, y=1.5cm]

\node (char1) {夏};
\node (char2) [right=1cm of char1]{天};
\node (char3) [right=1cm of char2]{太};
\node (char4) [right=1cm of char3]{热};

\node (eng2) [above=0.2cm of char3]{(too)};
\node (eng3) [above=0.2cm of char4]{(hot)};
\node (eng1) [left=1cm of eng2]{(summer)};

\foreach \x in {1,...,4}
	\node [circle, draw, minimum size=0.8cm] (emb1-\x) [below=0.2cm of char\x]{};
\foreach \x in {1,...,4}	
	\node [circle, draw, minimum size=0.8cm] (emb2-\x) [below=0.05cm of emb1-\x]{};
\foreach \x in {1,...,4}	
	\node [circle, draw, minimum size=0.8cm] (emb3-\x) [below=0.05cm of emb2-\x]{};
\foreach \x in {1,...,4}	
	\node[fit=(emb1-\x)(emb3-\x),inner sep=0, draw](emb-\x) {};

\node (note1) [left=0.3cm of emb2-1, text width=2cm, align=center]{3-gram character \\ representations};

\foreach \x in {1,...,4}
	\node [circle, draw, minimum size=0.8cm] (fgru-\x) [below=1cm of emb-\x]{GRU};	

\coordinate  [draw=none, left=0.6cm of fgru-1](fgru-0) ;
\coordinate  [draw=none, right=0.6cm of fgru-4](fgru-5);

\foreach \x [count=\xi from 1] in {0,...,4}
	\draw [-{Latex[length=2mm]}] (fgru-\x) -- (fgru-\xi);
 
\foreach \x in {1,...,4}
	\draw [-{Latex[length=2mm]}] (emb3-\x) -- (fgru-\x)[dashed];

\node (note2) [left=0.3cm of fgru-1, text width=2cm, align=center]{forward\\ RNN};

\foreach \x in {1,...,4}
	\node [circle, draw, minimum size=0.8cm] (bgru-\x) [below=1cm of fgru-\x]{GRU};

\coordinate  [draw=none, left=0.6cm of bgru-1](bgru-0) ;
\coordinate  [draw=none, right=0.6cm of bgru-4](bgru-5);

\foreach \x [count=\xi from 1] in {0,...,4}
	\draw [{Latex[length=2mm]}-] (bgru-\x) -- (bgru-\xi);

\foreach \x in {1,...,4}
	\draw[-{Latex[length=2mm]}] (emb3-\x.south) to [out=210,in=135] (bgru-\x.north)[dashed] ;

\node (note3) [left=0.3cm of bgru-1, text width=2cm, align=center]{backward\\ RNN};

\node [circle, draw, minimum size=1.2cm] (crf-1) [below=1cm of bgru-1]{B};
\node [circle, draw, minimum size=1.2cm] (crf-2) [below=1cm of bgru-2]{E};
\node [circle, draw, minimum size=1.2cm] (crf-3) [below=1cm of bgru-3]{S};
\node [circle, draw, minimum size=1.2cm] (crf-4) [below=1cm of bgru-4]{S};

\node (note4) [left=0.3cm of crf-1, text width=2cm, align=center]{CRF\\ Layer};

\foreach \x in {1,...,4}
	\draw [-{Latex[length=2mm]}] (bgru-\x) -- (crf-\x)[dashed];

\foreach \x in {1,...,4}
	\draw[-{Latex[length=2mm]}] (fgru-\x.south) to [out=335,in=40] (crf-\x.north)[dashed] ;

\foreach \x [count=\xi from 2] in {1,...,3}
	\draw [-] (crf-\x) -- (crf-\xi);

\node (post-3) [below=0.6cm of crf-3]{太};
\node (post-4) [below=0.6cm of crf-4]{热};
\node (post-12) [left=1cm of post-3]{夏天};

\node (note4) [left=1.7cm of post-12, text width=2cm, align=center]{Output};

\end{tikzpicture}
}
\caption{The BiRNN-CRF model for segmentation. The dashed arrows indicate that dropout layers are applied.}\label{fig:1}
\end{figure}
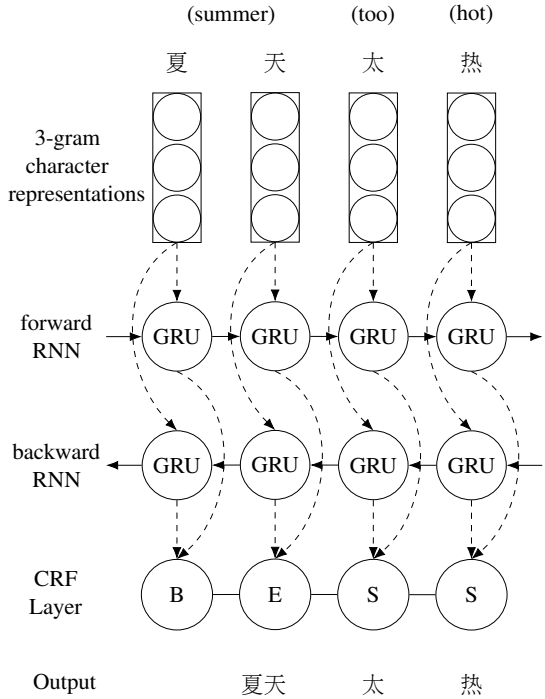

Our baseline model is shown in Figure \ref{fig:1}. We adopt the concatenated 3-gram model introduced in \newcite{shao17} as the vector representation of input characters. The pivot character in a given context is represented as the concatenation of the context-free vector along with the local bigram and trigram vectors. All the vectors are initialised randomly and separately. Utilising the concatenated n-grams ensures that the same character has different yet closely related vector representations in different contexts, which is an effective way to encode contextual features. We use a single vector to represent all the characters that appeared only once in the training set while training. This vector is later used as the representation for unknown characters in the development and test sets. The same representation scheme is also applied to local bigrams and trigrams. 

The character vectors are passed to the forward and backward recurrent layers. Gated recurrent units (GRU) \cite{cho2014properties} are employed as the basic recurrent cell to capture long term dependencies and global information. Compared to the more prevalent long-short term memory cells (LSTM) \cite{hochreiter1997long}, GRU has similar functionalities but fewer parameters \cite{chung2014empirical}. Dropout \cite{srivastava2014dropout} is applied to the input vectors as character representations and the outputs of the bidirectional recurrent layers. A first-order chain CRF layer is added on top of the recurrent layers to incorporate the transition information between consecutive tags, which ensures that the optimal sequence of tags over the entire sentence is obtained. The optimal sequence can be obtained efficiently via the Viterbi algorithm both for training and decoding. The time complexity is linear with respect to sentence length.

For the chain CRF interface while decoding, the final sequence of the position tags $y$ is obtained via the conditional scores $S(y_i|x_i)$ and the transition scores $T(y_i, y_j)$ given the input sequence $x$. In the baseline model, the optimal sequence is computed with respect to the scores returned by a single model:
 \begin{equation}
y^* = \argmax_{y \in L(x)} p(y|x; S, T)
\end{equation}
There is a post processing step to retrieve segmented units with respect to the predicted tags, which varies from different segmentation tasks and different formats of data sets.

\subsection{Tag Set}

\begin{figure}[t]
    \centering
    \scalebox{0.75}{
    \fbox{
        \begin{tabular}{rl}
        Characters: &\texttt{el\"{a}m\"{a} tuo kremppoja mukanaan .}\\
        Tags: &\texttt{BIIIEXBESXBIIIIIESSXBIIIIEBEXS}\\
        Segmented: & el\"{a}m\"{a} tu//o kremppo//j//a mukana//an .
        \end{tabular}
    }
    }
    \caption{Boundary tags employed for morpheme segmentation.}
    \label{fig:2}
\end{figure}

For word segmentation, we use four position tags B, I, E, and S to indicate a character positioned at the beginning (B), inside (I), or at the end (E) of a word, or occurring as a single-character word (S). We extend this tag set for morpheme segmentation by adding an extra tag X to represent the word boundaries. Figure \ref{fig:2} illustrates the input characters and the boundary tags to be predicted in morpheme segmentation.

Unlike previous work of \newcite{ruokolainen2013supervised} and \newcite{wang2016mor}, our model performs morpheme segmentation at the sentence level rather than the word level to incorporate information beyond word boundaries.  

\subsection{Ensemble Decoding}

We use a simple ensemble averaging technique to mitigate the deviations caused by random weight initialisation of the neural network and improve the baseline.   For ensemble decoding, both the conditional scores $S(y_i|x_i)$ and the transition scores $T(y_i, y_j)$ are averaged over four models with identical parameter settings but trained independently with different random seeds:
\begin{equation}
y^* = \argmax_{y \in L(x)} p(y|x; \bar{\{S\}}, \bar{\{T\}})
\end{equation}

\subsection{Implementation}

\begin{table}
\centering
\scalebox{0.90}{
\begin{tabular}{l|c}
\hline
Character vector size & 50 \\
2-gram and 3-gram vector sizes & 50 \\
\hline
GRU state size & 200 \\
\hline
Optimizer & Adagrad \\
Initial learning rate & 0.1 \\
Decay rate & 0.05 \\
Gradient Clipping & 5.0 \\
\hline
Dropout rate & 0.5 \\
Batch size & 10 \\
\hline
Length limit & 300 \\
\hline
\end{tabular}
}
\caption{Hyper-parameters for segmentation.}\label{tab:1}
\end{table}

Our neural segmenter is implemented using the TensorFlow 1.2.0 library \citep{abadi2016tensorflow}. The bucket model is applied so that the training and tagging speed of our neural network on GPU devices can be drastically improved. The training time is proportional to the size of the training set. We provide an open-source implementation of our method.\footnote{
https://github.com/yanshao9798/segmenter
}

Table \ref{tab:1} shows the adopted hyper-parameters. We use one set of parameters for both tasks on all the provided data sets. The weights of the neural networks, including the character vectors, are initialised using the scheme introduced in \newcite{glorot2010understanding}. The network is trained with the error back-propagation algorithm. The vector representations of input characters are fine-tuned during training by back-propagating gradients. Adagrad  \cite{duchi2011adaptive} with mini-batches is employed for optimisation with the initial learning rate $ \eta_0 = 0.1$, which is updated with a decay rate $\rho = 0.05$ as $\eta_t = \frac{\eta_0}{\rho (t - 1) + 1} $,  where $t$ is the index of the current epoch. To increase the efficiency and reduce memory demand both for training and decoding, we chop the sentences longer than 300 characters. For decoding, the chopped sentences are recovered after being processed.

The model is optimised according to its performance on the development sets. F1-score with respect to the basic segmented unit is employed to measure the performance of the model after each epoch during training. In our experiments, the models are trained for 30 epochs. To ensure that the weights are well optimised, we only adopt the best epoch after the model is trained at least for 5 epochs. 

\section{Experiments}

\subsection{Data Sets}

There are in total 10 data sets provided in the MLP 2017 shared tasks. Traditional Chinese, Japanese and Vietnamese are for the word segmentation task, while Basque, Farsi, Filipino, Finnish, Kazakh, Marathi and Uyghur are for morpheme segmentation. The provided languages vary substantially both in typology and written form. The sizes of the data sets are also different. The detailed information can be found in Table \ref{tab:2}.

In our experiments, we only use the official training data to build separate segmentation models for each language without utilising any external resources. For Vietnamese, we use the space-delimited units as the basic elements for boundary prediction. For all the rest, no language-specific modification or adjustment is made.   

\subsection{Experimental Results}

\begin{table*}[!htbp]
\centering
\begin{tabular}{c|ccc||ccc|ccc||ccc|r|r}
\hline
 & \multicolumn{3}{c||}{Size} & \multicolumn{3}{c|}{Baseline} & \multicolumn{3}{c||}{Ensemble} & Diff 1 & Diff 2\\
\hline
Dataset & Train & Dev & Test & P & R & F & P & R & F & F & F  \\
\hline
Chinese & 2,029 & 250 & 250 & 84.2 & 87.1 & 85.7 & 85.4 & 87.8 & 86.6 & +0.9 & +9.7  \\
Japanese & 1,600 & 200 & 200 & 96.1 & 97.8 & 96.9 & 96.6 & 97.7 & 97.2 & +0.3 & +3.4 \\
Vietnamese & 3,000 & 500 & 500 & 90.9 & 92.8 & 91.8 & 92.0 & 92.8 & 92.4 & +0.6 & -  \\
\hline
\hline
Basque & 599 & 100 & 100 & 81.5 & 77.2 & 79.3 & 82.4 & 80.8 & 81.6 & +2.3 & +28.2  \\
Farsi & 500 & 100 & 100 & 77.6 & 74.0 & 75.8 & 77.9 & 76.2 & 77.0 & +1.2 & +17.0  \\
Filipino & 1,999 & 200 & 200 & 91.2 & 93.0 & 92.1 & 92.0 & 92.4 & 92.2 & +0.1 & -  \\
Finnish & 3,537 & 750 & 762 & 89.8 & 90.5 & 90.2 & 90.9 & 90.5 & 90.7 & +0.5 & +26.5  \\
Kazakh & 7,298 & 999 & 1,000 & 97.0 & 97.1 & 97.1 & 97.5 & 97.5 & 97.5 & +0.4 & -  \\
Marathi & 5,098 & 450 & 450 & 95.1 & 93.3 & 94.2 & 95.1 & 93.8 & 94.4 & +0.2 & -  \\
Uyghur & 3,999 & 500 & 501 & 65.1 & 61.5 & 63.3 & 67.5 & 61.4 & 64.3 & +0.1 &  -0.8 \\
Uyghur* & 3,999 & 500 & 501 & 96.8 & 96.8 & 96.8 & 97.1 & 97.3 & 97.2 & +0.4 &  +32.7 \\
\hline
\end{tabular}
\caption{Data size in numbers of sentences and official evaluation results in precision (P), recall (R) and F1-score (F). Ensemble is in comparison to the single model as Baseline in Diff1. Baseline is in comparison to the best or second best systems in Diff 2. Asterisk indicates post-official evaluation runs.}\label{tab:2}
\end{table*}

For word segmentation, word level precision, recall and F1-score are employed as the evaluation metrics. For morpheme segmentation, precision, recall and F1-score are calculated only with respect to the identified prefixes and suffixes. 

The official experimental results are shown in Table \ref{tab:2}. The results of both the single and ensemble models are presented. The F1-scores of the single model as Baseline are in comparison to the ensemble model as well as the best performing systems among the other participants of the shared task. 

In general, the BiRNN-CRF model is effective for both word segmentation and morpheme segmentation. Our baseline model is substantially better than the rest of the participating systems on all the languages, especially for morpheme segmentation. Due to some encoding issues, our system is relatively under-performing on Uyghur referring to the official scores. We fixed the problem and report the corrected scores. 

The ensemble decoding is beneficial across all the data sets, but the overall improvement is rather marginal, especially if the baseline accuracy is very high. It is nonetheless helpful if the training sets are small as in the cases of Basque and Farsi. 

For word segmentation, we can see that very high accuracy is obtained on Japanese in spite of the relatively small training set. The writing system of Japanese is a combination of hiragana, katakana and Chinese characters (kanji). The switching of different types of characters can be an indicator for word boundaries in a sentence. As opposed to Japanese, Chinese and Vietnamese only contain one type of characters. The identification of word boundaries depends more heavily on the context. Additionally, the Chinese data set is composed of sentences from web search in different genres and the percentage of out-of-vocabulary words is high in the test set, which makes the segmentation task more challenging.

For morpheme segmentation, the size of the training set has the biggest impact on accuracy. The evaluation scores on Basque and Farsi are therefore drastically lower than the others. Unlike Chinese and Japanese characters and the space-delimited units in Vietnamese, individual characters by themselves in the languages for morpheme segmentation are less informative and the character vocabulary size is much smaller. The types of prefixes and suffixes to be identified are very limited and less ambiguous. As long as the data set is standardised and properly tokenised, very high accuracies can be achieved across languages in different writing systems given sufficient training data.

\section{Conclusions}

This paper presents our segmentation system for the MLP 2017 shared task on word segmentation and morpheme segmentation. Viewing both word and morpheme segmentation as character level sequence labelling tasks, we adapt the BiRNN-CRF model that has been applied to various sequence labelling tasks previously. Regardless of the vast variety of the data sets, we employ a universal model that uses a single set of hyper-parameters on all the languages without any task and language-specific adaptations. The evaluation results indicate that our model is effective for both segmentation tasks.

In general, the proposed model achieves relatively high accuracies across all the languages for both tasks if sufficient amounts of training data are provided. However, as both word segmentation and morpheme segmentation are at the very low levels of the complete natural language processing framework, the segmentation errors propagate further to higher level tasks. Thus, it is still very valuable to develop systems with higher performances in the future. In addition, we will explore the possibility of adapting the proposed model to low-resource languages using cross-lingual approaches if sufficient amount of training data is not available.

\section*{Acknowledgments}

We acknowledge the computational resources provided by CSC in Helsinki and Sigma2 in Oslo through NeIC-NLPL (www.nlpl.eu).

\bibliography{acl2017}
\bibliographystyle{acl_natbib}

\appendix

\end{CJK}

\end{document}